\documentclass[letterpaper, 10 pt, conference]{ieeeconf}  

\IEEEoverridecommandlockouts                              

\usepackage{float}
\usepackage{wrapfig}
\usepackage{setspace}
\usepackage{url}
\usepackage{graphicx} 
\usepackage{cite}
\usepackage{dcolumn,booktabs}
\usepackage{array}
\usepackage{multirow}
\usepackage{multicol}
\usepackage{hyperref}

\newcommand{\mc}[1]{#1}

\hypersetup{%
 setpagesize=false,%
 bookmarks=true,%
 bookmarksdepth=tocdepth,%
 bookmarksnumbered=true,%
 colorlinks=false,%
 pdftitle={},%
 pdfsubject={},%
 pdfauthor={},%
 pdfkeywords={}}

\title{\LARGE \bf JaywalkerVR: A VR System for Collecting Safety-Critical\\Pedestrian-Vehicle Interactions}

\author{Kenta Mukoya$^{1}$, Erica Weng$^{1}$, Rohan Choudhury$^{1}$ and Kris Kitani$^{1}$ 
\thanks{$^{1}$ Robotics Institute, Carnegie Mellon University, Pittsburgh, PA, USA. Contact: \texttt{kmukoya@andrew.cmu.edu}}%
}
\begin{document}

\maketitle

\begin{abstract}
Developing autonomous vehicles that can safely interact with pedestrians requires large amounts of pedestrian and vehicle data in order to learn accurate pedestrian-vehicle interaction models. However, gathering data that include crucial but rare scenarios - such as pedestrians jaywalking into heavy traffic - can be costly and unsafe to collect. We propose a virtual reality human-in-the-loop simulator, JaywalkerVR, to obtain vehicle-pedestrian interaction data to address these challenges. Our system enables efficient, affordable, and safe collection of long-tail pedestrian-vehicle interaction data. Using our proposed simulator, we create a high-quality dataset with vehicle-pedestrian interaction data from safety critical scenarios called CARLA-VR. The CARLA-VR dataset addresses the lack of long-tail data samples in commonly used real world autonomous driving datasets. We demonstrate that models trained with CARLA-VR improve displacement error and collision rate by 10.7\% and 4.9\%, respectively, and are more robust in rare vehicle-pedestrian scenarios.
\end{abstract}

\section{Introduction}
Safe autonomous vehicles require precise, multi-modal trajectory prediction systems, especially in highly interactive environments with pedestrians.
A major issue for such learning-based prediction or planning systems is the lack of data in complex and dangerous scenes, especially as data-hungry models like Transformers \cite{vaswani_attention_2017} have become the standard. Collecting data from such scenes is challenging from public roads and public datasets in particular lack such scenarios. Structured data collection, in which human subjects carry out long-tail behaviors, can be dangerous. For example, asking children to jaywalk across a busy road is unsafe.

Several methods have been proposed to collect synthetic data using virtual environments to compensate for this gap. For example, Gustavo~\cite{c2} proposes a real-time simulator with a steering controller to acquire driving data in interactive scenarios, and Junwei~\cite{c3} proposes to collect behavior and trajectory data of pedestrians using a keyboard controller. These methods pre-define interactive scenes of vehicles and pedestrians in the simulator to generate datasets. However, these systems suffer from a large sim-to-real gap as the subject uses a keyboard controller or joystick to control pedestrians while watching the screen. These controllers cannot accurately reproduce walking behaviors because of the restriction of control freedom. For example, behaviors like waiting for the right time to jaywalk while watching for oncoming vehicles are difficult to reproduce with such input devices due to a lack of head-yaw angle data.  Body tracking with virtual reality (VR) has been proposed to solve this issue ~\cite{c4}.
Since VR headsets have an immersive 360-degree field of view, tracking the headset allows the collection of head rotation and yaw data.

We introduce a human-in-the-loop pedestrian VR simulator for autonomous driving which can replicate real pedestrian behaviors and interactions called JaywalkerVR based on CARLA~\cite{c14}. 
Using our system, we generate a large, high-quality dataset of vehicle-pedestrian interactions called CARLA-VR. We then demonstrate the benefit of this data by training several prediction models with this data and find significant improvements, especially in highly interactive scenes.

To summarize, our key contributions are:
\begin{itemize}
    \item We develop a virtual reality-based autonomous driving simulator, JaywalkerVR, that can realistically simulate vehicle-pedestrian interaction in long-tail scenarios.
    \item We collect a high-quality vehicle-pedestrian interaction dataset, CARLA-VR, obtained from real human subjects using our proposed VR simulator.
    \item  We provide experimental results supporting the benefit of our new CARLA-VR dataset for improving trajectory prediction performance in long-tail pedestrian-vehicle interaction scenarios.
\end{itemize}

\begin{figure*}[h]
    \begin{tabular}{cc}
      \begin{minipage}[t]{0.48\hsize}
        \centering
        \includegraphics[width=\linewidth]{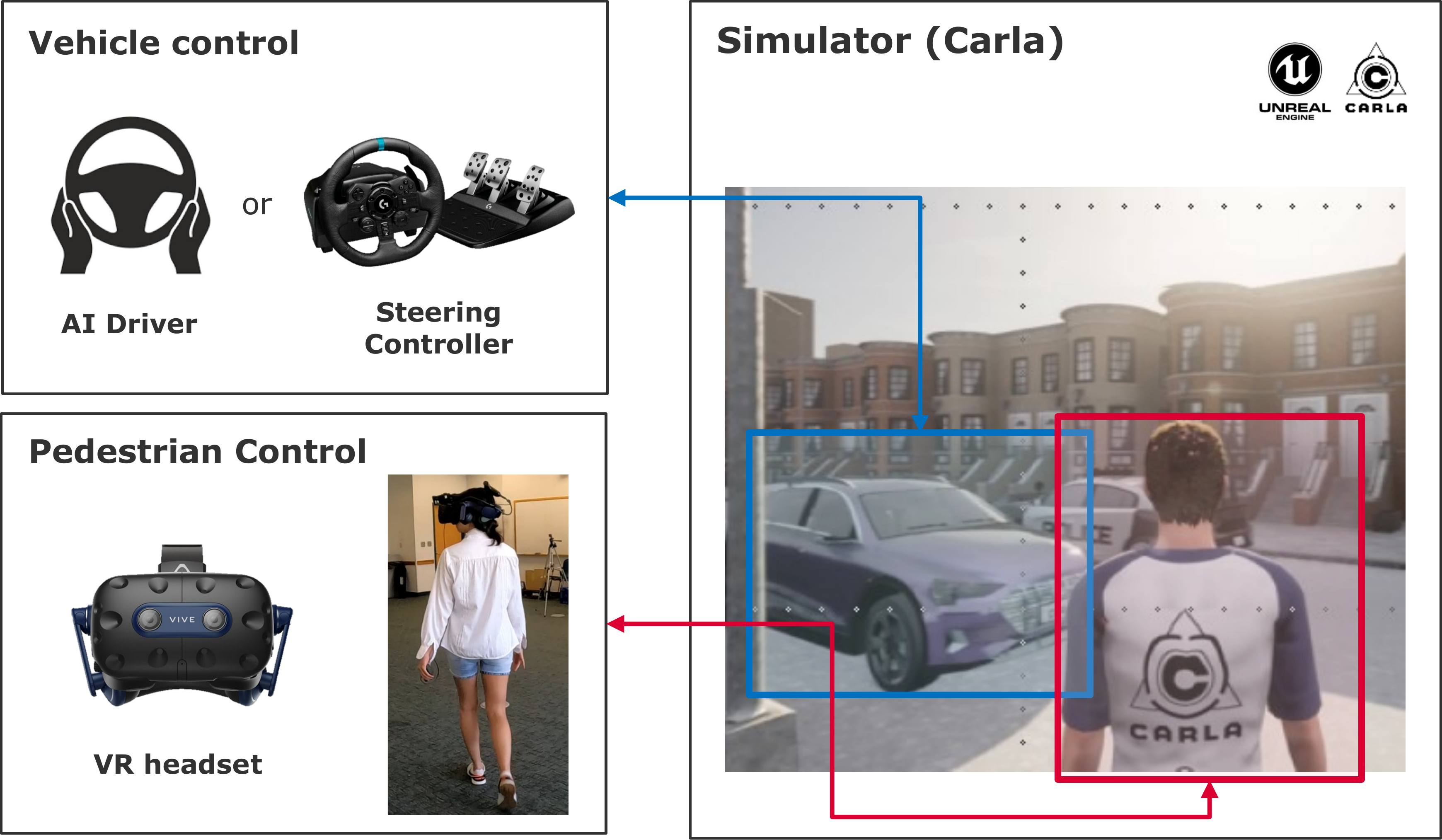}
        \caption{JaywalkerVR System Overview. People wearing VR headsets can experience 360-degree immersive simulator images and interact with vehicles and pedestrians in the same way they would in the real world. Simultaneously, the pedestrian avatar moves according to their movement in real world. Vehicles are controlled by CARLA AI agent (automatic control function) or manually using steering controller.}
    \label{fig:concept}
      \end{minipage} &
      \begin{minipage}[t]{0.48\hsize}
        \centering
        \includegraphics[width=\linewidth]{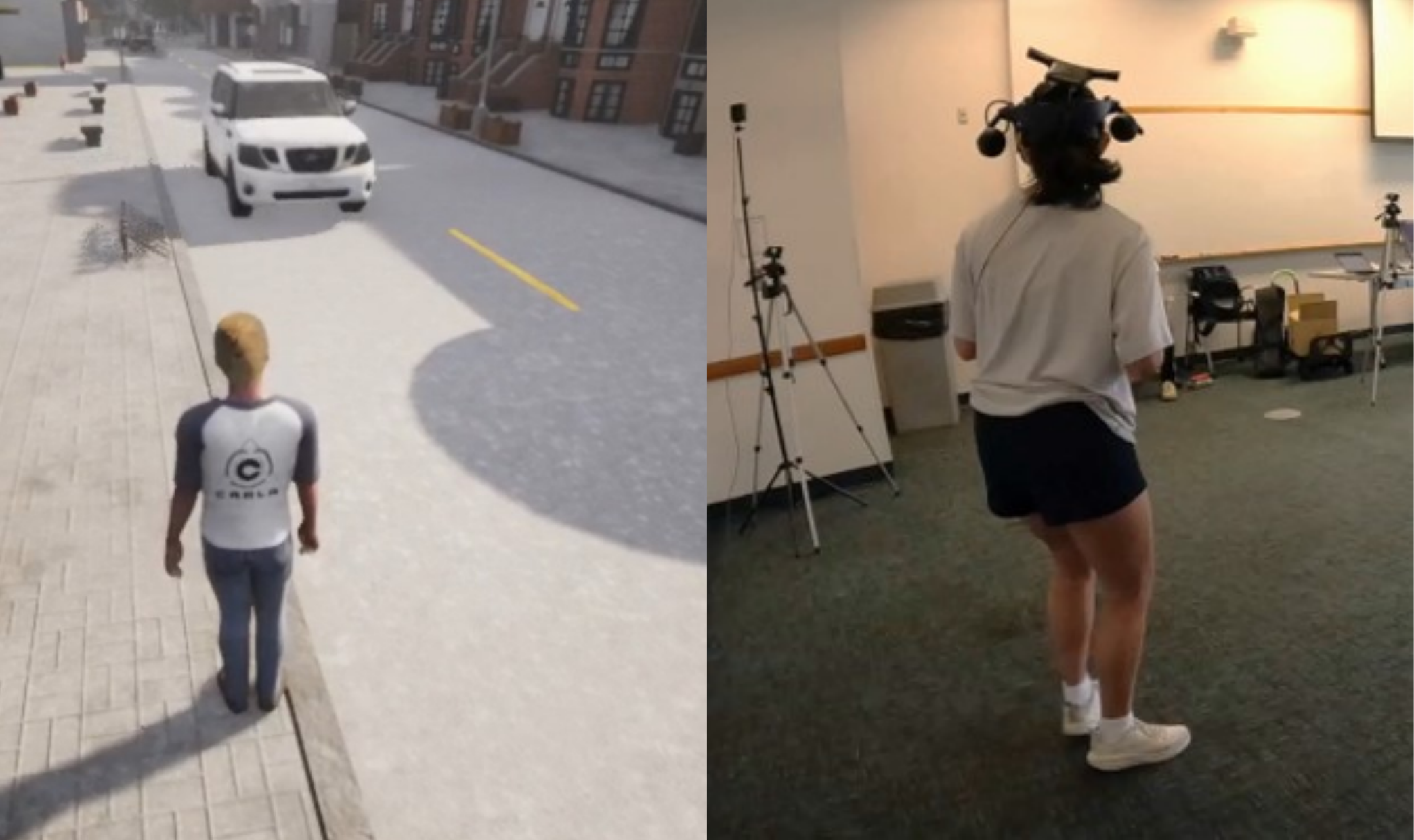}
        \caption{Example of JaywalkerVR simulation. Left: Subject's avatar in the JaywalkerVR from a third-person perspective. Right: Captured image of a subject wearing the VR headset in the data collection environment  in the real world.}
        \label{fig:VR demo}
      \end{minipage}
    \end{tabular}
  \end{figure*}

\section{Related Work}
In this section, we summarize the most relevant works related to a) trajectory prediction for autonomous driving, b) datasets for train/test trajectory prediction model, and c) autonomous driving simulators for simulating vehicle-pedestrian interaction.

\subsection{Trajectory Forecasting}
Modern trajectory forecasting models are deep, data-driven prediction models that predict futures for multiple interacting vehicles and pedestrians~\cite{c6,c7}. 
Some popular trajectory forecasting methods from recent years include 
methods built on deep generative architectures~\cite{kingma2013auto, goodfellow2014generative, rezende2015variational,gupta2018social,sadeghian2019sophie,kosaraju2019social,zhao2019multi,Kothari2022-cy,rhinehart2018r2p2,rhinehart2019precog,guan2020generative}, 
conditional variational autoencoders (CVAEs)~\cite{lee2017desire,yuan2019diverse,ivanovic2019trajectron,tang2019multiple,weng2020joint}, hierarchical architectures~\cite{Chiara2022-xa, mangalam_goals_2020, zhao_tnt_2020,chai_multipath_2019, chai2020multipath, Yue2022-os, Zhao2021-ca, Yao2021-xn, Wong2022-kx, liu_energy-based_2021, Xu2022-ki, liu_social_2021, liu_towards_2021}, 
and transformers~\cite{ivanovic2019trajectron, mangalam_it_2020, salzmann_trajectron_2021, kipf2018neural, alet_neural_2019,Liang2020-of, chen_relational_2020,su_trajectory_2022, mohamed_social-stgcnn_2020, Yu2020-df,Yuan2020-vw,Yuan2021-tp, Amirloo2022-se, Cao2022-uv, bahari_svg-net_2021,Huang2021-sv}.
Though there is much variety among architectures, one commonality they all share is that they rely on training on ample amounts of good quality data to produce accurate prediction results, 

\subsection{Trajectory Datasets}
Public datasets such as nuScenes~\cite{c10}, the Waymo Open Motion Dataset~\cite{womo}, Argoverse~\cite{chang_argoverse_2019}, and KITTI~\cite{Geiger2013-xv} are often used for training and testing of trajectory prediction models. 
These datasets are collected in the real world by real vehicles driving in public traffic environments. These datasets are dominated by commonly-occurring environments and scenes; there is little variety in available scenes, and there is a particular lack of uncommon environments such as narrow roads or alleyways, and uncommon scenarios such as pedestrian jaywalking, pedestrians walking alongside vehicles on the road, or dangerous or close contacts between pedestrians and vehicles. 
One method that is used to supplement real datasets is with more data from uncommon scenes is by generating synthetic data using traffic and pedestrian simulators~\cite{c12,liang_garden_2020-1,biswas_socnavbench_2021,kothari_human_2021}. 
With simulators, it is possible to generate data in many scenarios with low cost.  
However, in terms of collecting the pedestrian behavior data, most synthetic dataset generation methods use rudimentary autonomous policies~\cite{Van_den_Berg2011-yv,Guo2021-ec,Chen2016-xm} to generate pedestrian agent behavior. Other methods solicit input from real pedestrians via data-collection participants using mouse clicks or keyboard controls to control a pedestrian avatar in a virtual environment shown on a display screen~\cite{Liang2019-xo}.
These methods also have limitations, as clicks and keyboard controls fall short of the full degree of control pedestrians have over their movements and trajectories during navigation in real urban experiences.

\subsection{VR Pedestrian and Vehicle Simulators}
Some works have proposed using scenario simulators with VR headsets to collect pedestrian behavior data more accurate than that found in autonomous simulators or to study pedestrian responses to vehicle motion. 
For example, Dosovitskiy et al. (2017) and Schmitt et al. (2022) both created VR simulators where pedestrians are asked to click a button when they decide to cross the street in VR, and Silver et al. (2022) create VR driving simulators to record driver trajectory information ~\cite{Silvera2022-uk}. 

However, these works focus only on verifying pedestrian behavior rather than recording pedestrian trajectory and body pose data. To our knowledge, no prior work has collected data from human trajectories in uncommon but safety-critical scenarios, such as jaywalking, vulnerable road users, or pedestrian behavior in infrequently-seen urban environments.

\section{VR Simulator For Vehicle-Pedestrian Interaction Data Generation}

\begin{figure*}[h]
  \begin{center}
    \includegraphics[width=\linewidth]{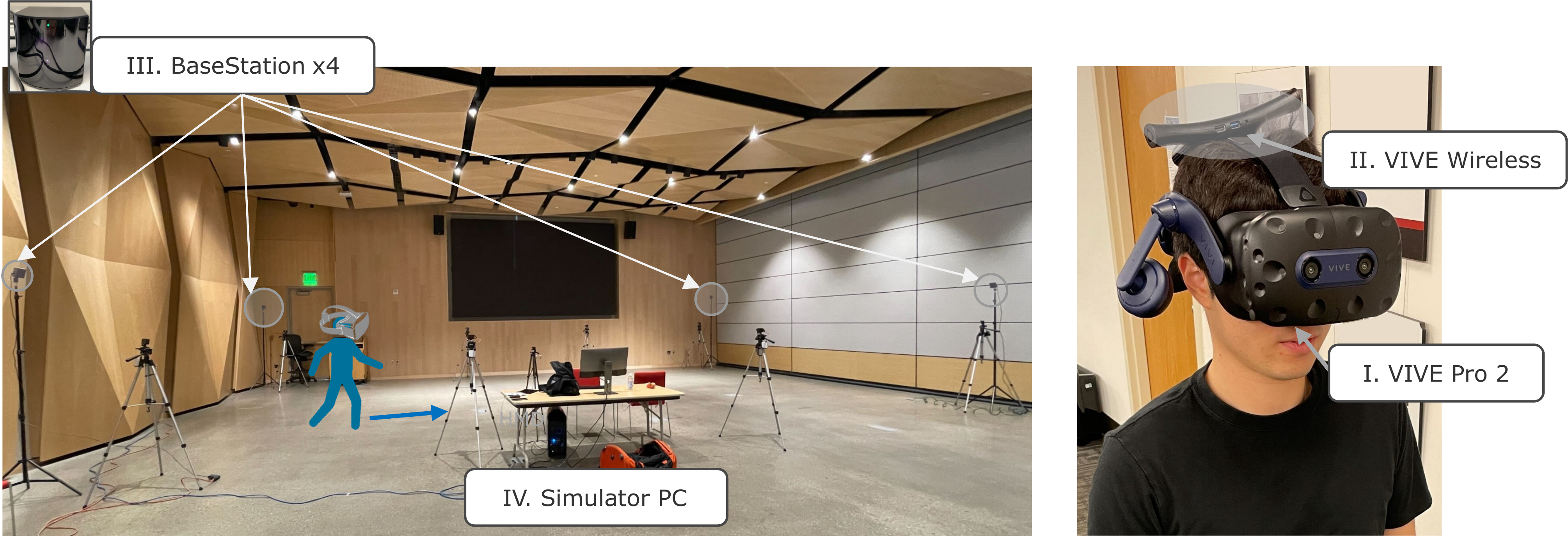}
    \caption{Room and equipment setting. (I) VIVE Pro 2: used for 1.visualizing simulator images and 2.tracking human position/rotation in the room (II) VIVE Wireless Adapter: used for allowing VIVE Pro 2 completely wireless (III) BaseStation 2.0: Track VR headset's position/rotation based on lighthouse tracking algorithm (IV) Simulator PC: Execute CARLA based VR simulator.}
    \label{fig:system setting}
  \end{center}
\end{figure*}

\subsection{System Overview} 
We developed our VR human-in-the-loop pedestrian simulator based on CARLA\cite{c16}, a popular open-source driving simulator for autonomous driving based on Unreal Engine 4 that includes convenient map and agent assets which we use for defining scenarios. \par
An overview of our simulator, JaywalkerVR, is shown in Fig.\ref{fig:concept}. We use a VR headset so that the human subjects interact with agents as realistically as possible compared to prior work. Since we need annotated interaction data between vehicles and pedestrians, especially pedestrian trajectory and head rotation data, our system simulates the walker avatar’s motion according to actual human motion.  To synchronize the motion between real human and pedestrian avatars in the simulation world, we use the tracking information from the headset such as 3D location and rotation angle.

\subsection{Walker Control $\&$ Tracking}
We use the tracking function of the VR headset to control the pedestrian avatar. This function relies on the HTC BaseStation 2.0, an ``Outside-In" tracking system which employs a lighthouse tracking method to accurately determine the position of the headset within the tracking range. The official tracking range extends up to approximately 10 meters in both dimensions, shown in Fig. \ref{fig:system setting}.

We synchronize the real-world sensor values of the headset and apply them to the entire skeleton mesh to obtain pedestrian positions then calibration the headset using the room size and position. Using this information, we use the SteamVR plugin in Unreal Engine to obtain the 3D position $[x, y, z]$ of VR headset and use it to control the position of pedestrian skeletal mesh in CARLA. In each scenario, we synchronize the pre-defined start position of the pedestrian avatar with the standing position of the human subject, and control the skeletal mesh model to follow the real human's movement.
We use the headset's yaw angle to adjust the yaw angle of the whole skeleton mesh. Finally, we update the pedestrian's movement animation to match their actual walking speed, enabling a person wearing a VR headset to control and move the avatar freely within the experimental setup.

\subsection{Pedestrian Model}
The walker skeleton model is provided in CARLA by default, and the movement of this skeleton model can be controlled by keyboard or joystick input devices. However, there are no native functions that control that skeleton model according to the movement of a VR headset. We modified the walker blueprint to control the skeleton model by synchronizing it with the motion the VR headset. The virtual camera module is attached to the walker's head, and the walker's blueprint is modified in order to get a first-person feel. 
The camera module acts as the avatar’s virtual eyes, and the skeletal mesh defines the walker's appearance.\par
In addition, we developed an IK setup (inverse kinematics) for the representation of walking animation. The skeleton model is designed to make walking motions in response to the movement speed of the VR headset. 

\subsection{Scenario Generation $\&$ Data Recording}
In order to define arbitrary scenarios for data collection, we implemented a scenario generation function using the CARLA Python API, in particular, the TrafficManager components. First, the CARLA AI Agent, which is the driving policy for autopilot implemented in the CARLA standard, was used to control the vehicle agent of CARLA, and the traffic flow was generated after the Autopilot function was enabled in each spawned vehicle. In terms of route planning, desired routes automatically run according to the route plan determined by the AI Agent by creating a route plan in which vehicle spawn points are arranged. In addition, the behavior of the AI Agent is used with the default setting and stops when a pedestrian is detected. Also, each agent's data, such as position and size, are collected at 20 Hz.

\subsection{Simulation Settings}
We used the HTC Vive Pro 2 VR headset which has SteamVR support. We used four HTC BaseStation 2.0 units for tracking the headset. We also installed the VIVE Wireless adapter, allowing the headset to be used completely wirelessly. We used a desktop PC which contains a PCI express slot to install the image emitter module of the VIVE Wireless adapter for the simulator, with an Intel core i9-12900KF CPU, NVIDIA GeForce RTX 3080 GPU, and 64GB RAM.

Since VIVE Wireless is only supported by Windows 10 or 11, we set up our CARLA-based VR pedestrian simulator on the Windows 11 desktop PC. We used  Unreal Engine UE 4.26.2 and CARLA 0.9.13.

\section{CARLA-VR Dataset}
\begin{figure*}[h]
  \begin{center}
    \includegraphics[width=0.8\linewidth]{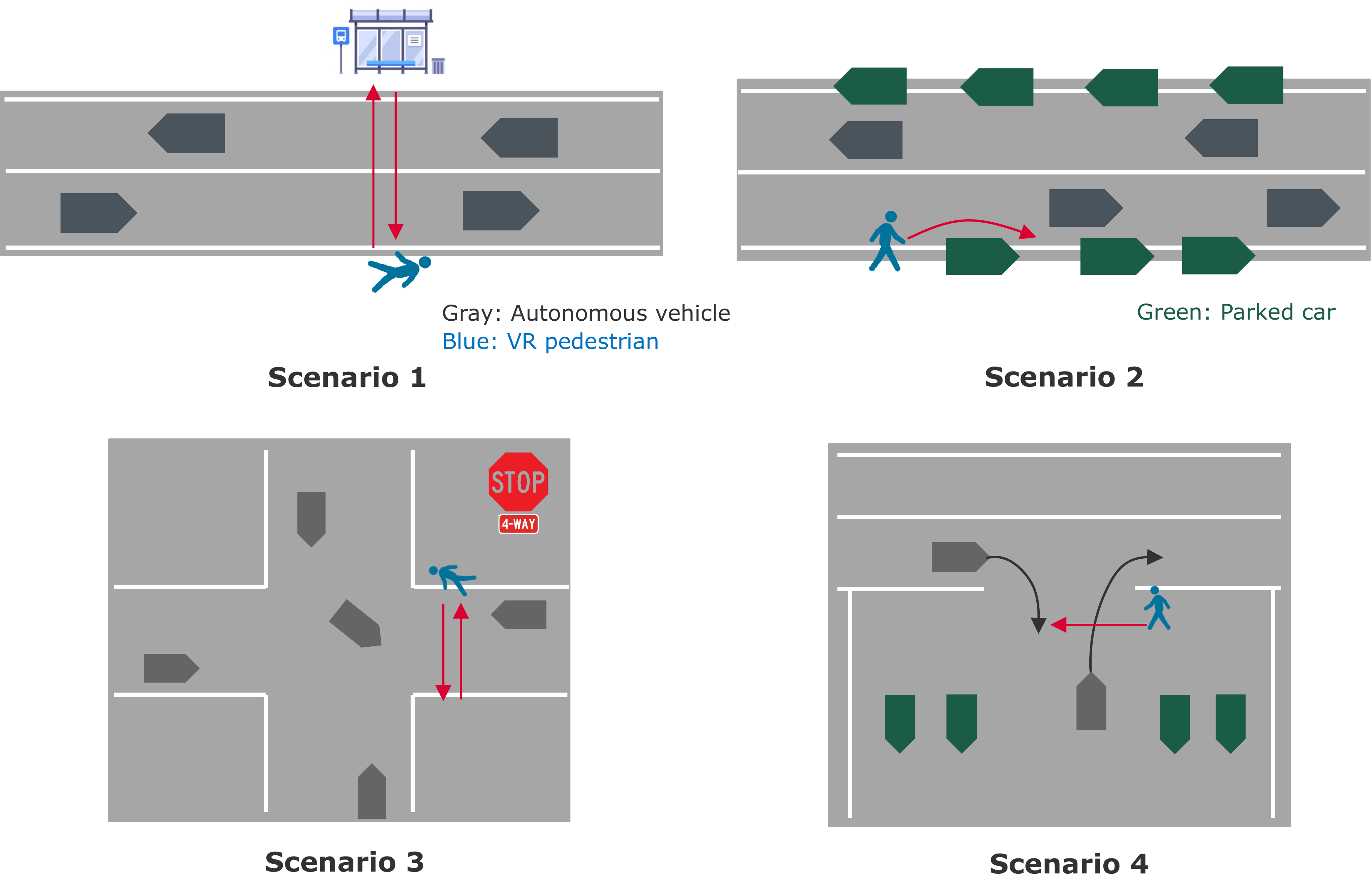}
    \caption{Experimental scenarios. (1) \textbf{Jaywalk:} Pedestrians jaywalk toward a bus stop while avoiding vehicles. (2) \textbf{Parked Cars:} Pedestrians avoid parked vehicles and move along the road. (3) \textbf{4-Way Stop:} Pedestrians cross the crosswalk, avoiding cars at a four-way stop. (4) \textbf{Parking Lot Entrance: }Pedestrians walk through the entrance or a parking lot paying attention to cars.}
    \label{fig:scenarios}
  \end{center}
\end{figure*}

\subsection{Interactive Scenarios}
We defined four pedestrian-vehicle interaction scenarios in CARLA for collecting VR human data as shown in Fig. \ref{fig:scenarios}.  
\vskip.5\baselineskip

\textbf{Jaywalk:}
Pedestrians jaywalk across a road while yielding to vehicles coming from both directions on a two-lane road. In this scenario, we expect the subjects to try to interact with oncoming vehicles, such as yielding to vehicles. And they cross the street on their own timing and with their own decision-making. For example, some subjects behave aggressively, but others will behave nervously and miss the opportunity to walk. Then, we obtain a variety of behaviors in each subject, such as different speeds of walking and different timings of crossing.
    
\textbf{Parked Cars:} 
Pedestrians walk along the edge of the road, avoiding parked vehicles and moving to a position one car ahead while paying attention to vehicles approaching from behind. In this scenario, we also expect subjects to start walking on their own timing.
    
\textbf{4-Way Stop:}
Pedestrians cross the crosswalk while paying attention to cars coming from four directions at a four-way stop. In this scenario, we expect subjects to cross the crosswalk at various times as decided by each of them for vehicles coming at them from different directions.

\textbf{Parking Lot Entrance:}
Pedestrians walk through the entrance to a parking lot while paying attention to and avoiding any entering and exiting vehicles. In this scenario, we expect subjects to behave by yielding or not yielding to the vehicles at various decisions.

\subsection{Dataset Details}
We collected data from 80 participants in each of the four scenarios. In the Jaywalk, Parked Cars and 4-Way Stop scenarios, the surrounding vehicles are controlled by a CARLA AI agent and in completely autonomous driving mode. In the Parking Lot Entrance scenario, the vehicles are controlled by a human driver using a steering controller, as CARLA did not support implementing a route plan for the vehicle to enter and exit the parking lot. We collected a total of 572 scenes comprising 12702 frames. The data contains position $[x,y,z] \mathrm{[m]}$, rotation $[\theta, \phi, \psi] \mathrm{[deg]}$, velocity $[v_x,v_y,v_z] \mathrm{[m/s]}$, acceleration $[a_x,a_y,a_z] \mathrm{[m/s^2]}$ in global coordinates in CARLA's map, object type (car, pedestrian) and object shape information [length, width, height]. Each scene data is between 10 and 30s long and was recorded at 20Hz. 

\section{Experiments}
We demonstrate that the realism of our data improves the performance of learned prediction models in long-tail interaction scenarios.

\subsection{Baseline model}
We use AgentFormer \cite{yuan_agentformer_2021} in all experiments for measuring trajectory forecasting performance. 
AgentFormer is a Transformer-based model that jointly models the time and social dimensions with an agent-aware attention mechanism. The model leverages a sequence representation of multi-agent trajectories by flattening trajectory features across time and agents and using the resulting spatiotemporal attention-based features for trajectory prediction. More details, such as the model architecture and training setup, are available in the original paper.
In our experiments, we generate 10 sample 2D trajectories for each agent by using past trajectories, yaw angle information, and a semantic segmentation image of a bird’s eye view obtained from CARLA as inputs. 

\newcolumntype{D}{D{.}{.}{4.5}}

\begin{table*}
    \caption{Evaluation Result}
    \label{table:evaluation_result}
    \centering
    \begin{tabular}{llp{1.5cm}lp{1.5cm}ll}
        \toprule
        Test Dataset & Model & \multicolumn{2}{c}{Marginal XDE$\mathrm{[m]}$ (K=10)}  & \multicolumn{2}{c}{Joint XDE$\mathrm{[m]}$ (K=10)} & \multicolumn{1}{c}{Collision rate$\mathrm{[-]}$}\\
        \cmidrule(lr){3-4} \cmidrule(l){5-6} \cmidrule(l){7-7}
        & & \mc{ADE$\downarrow$} & \mc{FDE$\downarrow$} & \mc{JADE$\downarrow$} & \mc{JFDE$\downarrow$}  & \mc{CR mean$\downarrow$} \\
        \midrule
        \multirow{2}{*}{nuScenes-prediction}            & AgentFormer-B{*}  & \textbf{1.2299} & \textbf{2.7175} & \textbf{2.4023} & \textbf{5.9062} & 0.1275 \\
        \addlinespace
                                                        & AgentFormer-VR{**} & 1.4408 & 3.1088 & 2.7020 & 6.5288 & \textbf{0.1186} \\
        \midrule
        \multirow{2}{*}{CARLA-VR}                       & AgentFormer-B{*}  & 1.1404 & 2.7243 & 1.9274 & 5.1474 & 0.3266 \\
        \addlinespace
        
                                                        & AgentFormer-VR{**} & \textbf{0.9319} & \textbf{2.1491} & \textbf{1.6193} & \textbf{4.1201} & \textbf{0.2856} \\
        \midrule
        \multirow{2}{*}{nuScenes-interaction}           & AgentFormer-B{*}  & 1.2712 & 2.8285 & 2.5995 & 6.4676 & 0.3170 \\
        \addlinespace
                                                        & AgentFormer-VR{**} & \textbf{1.1349} & \textbf{2.4637} & \textbf{2.2680} & \textbf{5.3770} & \textbf{0.3016} \\
        \bottomrule
        \multicolumn{5}{l}{\footnotesize ${*}$ AgentFormer trained on the nuScenes prediction dataset only}\\
        \multicolumn{5}{l}{\footnotesize ${**}$ AgentFormer trained on the nuScenes prediction dataset and the CARLA-VR dataset}\\
    \end{tabular}
\end{table*}

\subsection{Datasets}  
We use the following datasets in our experiments:  
\vskip.5\baselineskip

\textbf{nuScenes:}
nuScenes is a widely used public autonomous driving dataset with annotated data, such as position in global coordinates in nuScenes's map, rotation, and bounding box size at 2Hz. nuScenes also provides HD semantic maps with 11 semantic classes.

\textbf{nuScenes-prediction:}
We extract the nuScenes prediction dataset from annotated data for the nuScenes prediction challenge. This is used for pre-training of the trajectory prediction model and also for evaluation of prediction performance in the general scenes.

\textbf{nuScenes-interaction:}
To check the prediction model's performance in rare scenes, we extract interactive scenes from similar situations to our simulation scenarios (e.g. jaywalking) from annotated data on the nuScenes dataset, following the filtering method of \cite{c5}.  Since this dataset contains only vehicle-pedestrian interaction data that actually occurred in the real world, testing the prediction model with this dataset allows us to evaluate the model's performance in real-world interactive scenes. This dataset is used for the evaluation of prediction performance in interactive scenes in the real world.
\vskip.5\baselineskip
\textbf{CARLA-VR dataset:}
Our collected dataset contains rare vehicle-pedestrian interactive scene data from the VR simulator described in Section IV. We use it for pre-training of the trajectory prediction model and also for evaluation of prediction performance in the interactive scenes in the simulator world. To align the sampling rate, CARLA-VR dataset is also resampled from 20Hz to 2Hz. \par

\subsection{Experiment settings and evaluation metrics}
Our baseline is state-of-the-art AgentFormer trained on the nuScenes prediction dataset, denoted \textbf{AgentFormer-B}. To demonstrate the utility of our proposed dataset, we further train AgentFormer-B on CARLA-VR to get \textbf{AgentFormer-VR}.
We then evaluate both models' performance on nuScenes-prediction, CARLA-VR interaction, and nuScenes-interaction.
We use the following metrics to measure performance: 
\vskip.5\baselineskip
\textbf{Marginal XDE:} XDE encompasses Marginal Average Displacement Error (ADE) and Marginal Final Displacement Error (FDE), and these are commonly used for evaluating how the predicted trajectory is close to ground truth(GT) trajectory. Since AgentFormer generates 10 sample trajectory sets, we evaluate minXDE, the top-K minimum error. \par
\textbf{Joint XDE:}  Unlike XDE, Joint XDE(JXDE) evaluates scene-level ADE/FDE\cite{c17}. Since this metrics calculate the average error over all agents within a sample before we select the best one, we cannot mix-and-match agents between different samples. This means we can evaluate how the prediction result (top-K sample) is close to GT trajectory with considering social-interaction at scene-level. Same as XDE, we evaluate minJXDE (top-K minimum error). \par
\textbf{Collision Rate:} Collision Rate (CR) evaluates whether the predicted trajectories of each agent collide with each other within the same prediction timestep. 

\begin{figure*}[h]
    \centering
    \includegraphics[width=0.85\linewidth]{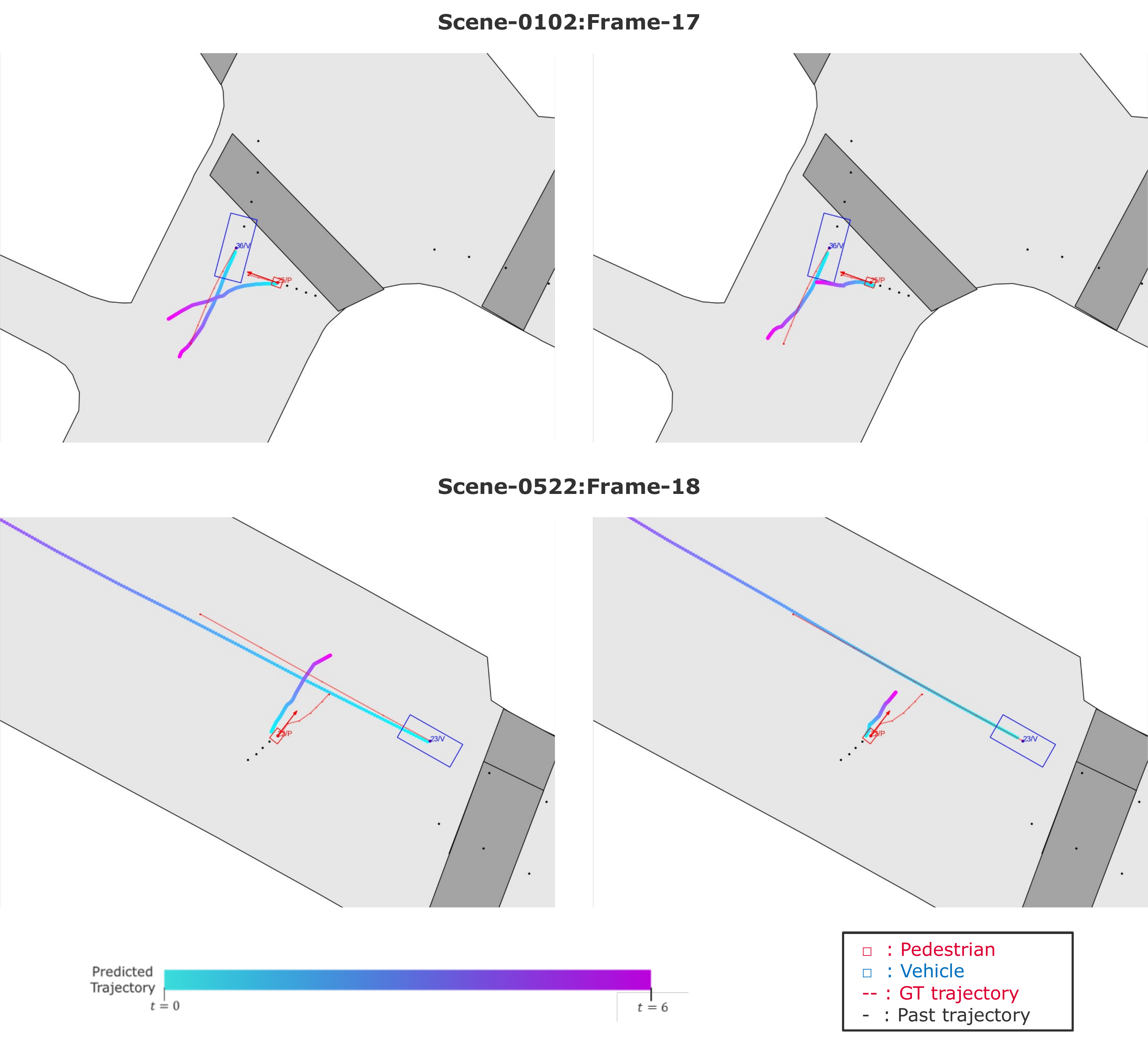}
    \caption{Visualization result. Left figure: predicted trajectories of AgentFormer-B, Right figure: predicted trajectories of AgentFormer-VR. Light gray areas indicate the drivable area and dark gray areas indicate pedestrian crossings.}
    \label{fig:vis result}
\end{figure*}
 
\subsection{Experimental results}
The results of the experiments are listed in Table. \ref{table:evaluation_result}. In terms of the evaluation result of CARLA-VR dataset and nuScenes interaction dataset,  all metrics improve when incorporating our CARLA-VR dataset. Marginal XDE performance improves by 10.7-12.8\%, and Joint XDE also improves by 12.6-16.9\%. Further, the most important metric for safety - collision rate - improves by 4.9\%. \par
In Fig. \ref{fig:vis result}, we show predicted trajectories from AgentFormer-B (left) and AgentFormer-VR (right). The GT trajectories are drawn in red, and the best predicted trajectories are shown with time-varying color. We find that AgentFormer-B, only trained on nuScenes prediction dataset, often predicts trajectories for pedestrians that lead them into direct collision with vehicles. We attribute this to the rarity of dangerous pedestrian-vehicle interactions in the real-world nuScenes dataset. On the other hand, when AgentFormer leverages our safety-critical interaction dataset, we see in the right figure that the pedestrian is predicted to \textit{yield} to the incoming vehicle, better matching the ground truth trajectory. These qualitative visualizations corroborate our quantitative results that the proposed CARLA-VR dataset, containing safety-critical pedestrian-vehicle interactions, better enables trajectory prediction models to model agent behavior in dangerous and rare scenarios.\par

\subsection{Discussion}

Our results show that the prediction model becomes more robust in real-world interactive scenes through fine-tuning on the CARLA-VR dataset. In particular, minJXDE and CR decreases substantially for nuScenes-interaction - the most safety-critical and difficult scenarios in the nuScenes dataset. Furthermore, AgentFormer-VR improves collision rates across all datasets. This is particularly crucial in evaluating trajectory forecasting models, as the ability to predict plausible trajectories with minimal collisions important for autonomous driving applications. While performance in the minJXDE metric drops for the nuScenes-prediction test set, we emphasize that that the full nuScenes dataset mostly consists of common or simpler driving scenarios \cite{womo}, and that evaluation on the more complex and interactive driving subset - nuScenes-interaction - is more critical. For these more safety-critical and dynamic scenarios, leveraging our CARLA-VR dataset substantially improves the robustness of interaction-aware motion predictions.


\section{Conclusions}
We present JaywalkerVR, a human-in-the-loop VR pedestrian simulator enabling the collection of realistic long-tail vehicle-pedestrian interaction scenario data. We also present the new CARLA-VR dataset, which contains rich, interactive vehicle-pedestrian scenario data from actual humans. In particular, our use of VR in data collection enables accurate trajectory and head angle annotations.  Finally, we demonstrate the effectiveness of this dataset for training trajectory forecasting models. Fine-tuning on the CARLA-VR dataset improved XDE, JXDE and CR, especially in highly interactive scenes. Our experiments show that our dataset and data collection pipeline will be effective tools for developing more robust prediction algorithms moving forward.

\section*{ACKNOWLEDGMENTS}
This research was supported in part by DENSO International America, Inc., the Ford Fellowship and the NSF Graduate Research Fellowship Program.

\bibliographystyle{IEEEtran}
\bibliography{paperpile, other}

\begin{thebibliography}{10}
\providecommand{\url}[1]{#1}
\csname url@rmstyle\endcsname
\providecommand{\newblock}{\relax}
\providecommand{\bibinfo}[2]{#2}
\providecommand\BIBentrySTDinterwordspacing{\spaceskip=0pt\relax}
\providecommand\BIBentryALTinterwordstretchfactor{4}
\providecommand\BIBentryALTinterwordspacing{\spaceskip=\fontdimen2\font plus
\BIBentryALTinterwordstretchfactor\fontdimen3\font minus \fontdimen4\font\relax}
\providecommand\BIBforeignlanguage[2]{{%
\expandafter\ifx\csname l@#1\endcsname\relax
\typeout{** WARNING: IEEEtran.bst: No hyphenation pattern has been}%
\typeout{** loaded for the language `#1'. Using the pattern for}%
\typeout{** the default language instead.}%
\else
\language=\csname l@#1\endcsname
\fi
#2}}

\bibitem{c2}
\BIBentryALTinterwordspacing
G.~Silvera, A.~Biswas, and H.~Admoni, ``Dreyevr: Democratizing virtual reality driving simulation for behavioural {\&} interaction research,'' \emph{CoRR}, vol. abs/2201.01931, 2022. [Online]. Available: \url{https://arxiv.org/abs/2201.01931}
\BIBentrySTDinterwordspacing

\bibitem{c3}
\BIBentryALTinterwordspacing
J.~Liang, L.~Jiang, K.~P. Murphy, T.~Yu, and A.~G. Hauptmann, ``The garden of forking paths: Towards multi-future trajectory prediction,'' \emph{CoRR}, vol. abs/1912.06445, 2019. [Online]. Available: \url{http://arxiv.org/abs/1912.06445}
\BIBentrySTDinterwordspacing

\bibitem{c4}
T.~T.~M. Tran, C.~Parker, and M.~Tomitsch, ``A review of virtual reality studies on autonomous vehicle–pedestrian interaction,'' \emph{IEEE Transactions on Human-Machine Systems}, vol.~51, no.~6, pp. 641--652, 2021.

\bibitem{c14}
\BIBentryALTinterwordspacing
A.~Dosovitskiy, G.~Ros, F.~Codevilla, A.~M. L{\'{o}}pez, and V.~Koltun, ``{CARLA:} an open urban driving simulator,'' \emph{CoRR}, vol. abs/1711.03938, 2017. [Online]. Available: \url{http://arxiv.org/abs/1711.03938}
\BIBentrySTDinterwordspacing

\bibitem{c6}
R.~Huang, H.~Xue, M.~Pagnucco, F.~Salim, and Y.~Song, ``Multimodal trajectory prediction: A survey,'' 2023.

\bibitem{c7}
Y.~Huang, J.~Du, Z.~Yang, Z.~Zhou, L.~Zhang, and H.~Chen, ``A survey on trajectory-prediction methods for autonomous driving,'' \emph{IEEE Transactions on Intelligent Vehicles}, vol.~7, no.~3, pp. 652--674, 2022.

\bibitem{kingma2013auto}
D.~P. Kingma and M.~Welling, ``Auto-encoding variational bayes,'' \emph{arXiv preprint arXiv:1312.6114}, 2013.

\bibitem{goodfellow2014generative}
I.~J. Goodfellow, J.~Pouget-Abadie, M.~Mirza, B.~Xu, D.~Warde-Farley, S.~Ozair, A.~Courville, and Y.~Bengio, ``Generative adversarial networks,'' \emph{arXiv preprint arXiv:1406.2661}, 2014.

\bibitem{rezende2015variational}
D.~Rezende and S.~Mohamed, ``Variational inference with normalizing flows,'' in \emph{International Conference on Machine Learning}.\hskip 1em plus 0.5em minus 0.4em\relax PMLR, 2015, pp. 1530--1538.

\bibitem{gupta2018social}
A.~Gupta, J.~Johnson, L.~Fei-Fei, S.~Savarese, and A.~Alahi, ``Social gan: Socially acceptable trajectories with generative adversarial networks,'' in \emph{Proceedings of the IEEE Conference on Computer Vision and Pattern Recognition}, 2018, pp. 2255--2264.

\bibitem{sadeghian2019sophie}
A.~Sadeghian, V.~Kosaraju, A.~Sadeghian, N.~Hirose, H.~Rezatofighi, and S.~Savarese, ``Sophie: An attentive gan for predicting paths compliant to social and physical constraints,'' in \emph{Proceedings of the IEEE/CVF Conference on Computer Vision and Pattern Recognition}, 2019, pp. 1349--1358.

\bibitem{kosaraju2019social}
V.~Kosaraju, A.~Sadeghian, R.~Mart{\'\i}n-Mart{\'\i}n, I.~D. Reid, H.~Rezatofighi, and S.~Savarese, ``Social-bigat: multimodal trajectory forecasting using bicycle-gan and graph attention networks,'' in \emph{Advances in Neural Information Processing Systems 2019}.\hskip 1em plus 0.5em minus 0.4em\relax Neural Information Processing Systems (NIPS), 2019.

\bibitem{zhao2019multi}
T.~Zhao, Y.~Xu, M.~Monfort, W.~Choi, C.~Baker, Y.~Zhao, Y.~Wang, and Y.~N. Wu, ``Multi-agent tensor fusion for contextual trajectory prediction,'' in \emph{Proceedings of the IEEE/CVF Conference on Computer Vision and Pattern Recognition}, 2019, pp. 12\,126--12\,134.

\bibitem{Kothari2022-cy}
P.~Kothari and A.~Alahi, ``Safety-compliant generative adversarial networks for human trajectory forecasting,'' Sept. 2022.

\bibitem{rhinehart2018r2p2}
N.~Rhinehart, K.~M. Kitani, and P.~Vernaza, ``R2p2: A reparameterized pushforward policy for diverse, precise generative path forecasting,'' in \emph{Proceedings of the European Conference on Computer Vision (ECCV)}, 2018, pp. 772--788.

\bibitem{rhinehart2019precog}
N.~Rhinehart, R.~McAllister, K.~Kitani, and S.~Levine, ``Precog: Prediction conditioned on goals in visual multi-agent settings,'' in \emph{Proceedings of the IEEE/CVF International Conference on Computer Vision}, 2019, pp. 2821--2830.

\bibitem{guan2020generative}
J.~Guan, Y.~Yuan, K.~M. Kitani, and N.~Rhinehart, ``Generative hybrid representations for activity forecasting with no-regret learning,'' in \emph{Proceedings of the IEEE/CVF Conference on Computer Vision and Pattern Recognition}, 2020, pp. 173--182.

\bibitem{lee2017desire}
N.~Lee, W.~Choi, P.~Vernaza, C.~B. Choy, P.~H. Torr, and M.~Chandraker, ``Desire: Distant future prediction in dynamic scenes with interacting agents,'' in \emph{Proceedings of the IEEE Conference on Computer Vision and Pattern Recognition}, 2017, pp. 336--345.

\bibitem{yuan2019diverse}
Y.~Yuan and K.~Kitani, ``Diverse trajectory forecasting with determinantal point processes,'' \emph{arXiv preprint arXiv:1907.04967}, 2019.

\bibitem{ivanovic2019trajectron}
B.~Ivanovic and M.~Pavone, ``The trajectron: Probabilistic multi-agent trajectory modeling with dynamic spatiotemporal graphs,'' in \emph{Proceedings of the IEEE/CVF International Conference on Computer Vision}, 2019, pp. 2375--2384.

\bibitem{tang2019multiple}
Y.~C. Tang and R.~Salakhutdinov, ``Multiple futures prediction,'' \emph{arXiv preprint arXiv:1911.00997}, 2019.

\bibitem{weng2020joint}
X.~Weng, Y.~Yuan, and K.~Kitani, ``Joint 3d tracking and forecasting with graph neural network and diversity sampling,'' \emph{arXiv preprint arXiv:2003.07847}, 2020.

\bibitem{Chiara2022-xa}
L.~F. Chiara, P.~Coscia, S.~Das, S.~Calderara, R.~Cucchiara, and L.~Ballan, ``Goal-driven self-attentive recurrent networks for trajectory prediction,'' pp. 2518--2527, Apr. 2022.

\bibitem{mangalam_goals_2020}
\BIBentryALTinterwordspacing
K.~Mangalam, Y.~An, H.~Girase, and J.~Malik, ``From {Goals}, {Waypoints} \& {Paths} {To} {Long} {Term} {Human} {Trajectory} {Forecasting},'' \emph{arXiv:2012.01526 [cs]}, Dec. 2020, arXiv: 2012.01526. [Online]. Available: \url{http://arxiv.org/abs/2012.01526}
\BIBentrySTDinterwordspacing

\bibitem{zhao_tnt_2020}
\BIBentryALTinterwordspacing
H.~Zhao, J.~Gao, T.~Lan, C.~Sun, B.~Sapp, B.~Varadarajan, Y.~Shen, Y.~Shen, Y.~Chai, C.~Schmid, C.~Li, and D.~Anguelov, ``{TNT}: {Target}-{driveN} {Trajectory} {Prediction},'' \emph{arXiv:2008.08294 [cs]}, Aug. 2020, arXiv: 2008.08294. [Online]. Available: \url{http://arxiv.org/abs/2008.08294}
\BIBentrySTDinterwordspacing

\bibitem{chai_multipath_2019}
\BIBentryALTinterwordspacing
Y.~Chai, B.~Sapp, M.~Bansal, and D.~Anguelov, ``{MultiPath}: {Multiple} {Probabilistic} {Anchor} {Trajectory} {Hypotheses} for {Behavior} {Prediction},'' \emph{arXiv:1910.05449 [cs, stat]}, Oct. 2019, arXiv: 1910.05449. [Online]. Available: \url{http://arxiv.org/abs/1910.05449}
\BIBentrySTDinterwordspacing

\bibitem{chai2020multipath}
\BIBentryALTinterwordspacing
Y.~Chai, B.~Sapp, M.~Bansal, and D.~Anguelov, ``Multipath: Multiple probabilistic anchor trajectory hypotheses for behavior prediction,'' in \emph{Proceedings of the Conference on Robot Learning}, ser. Proceedings of Machine Learning Research, vol. 100.\hskip 1em plus 0.5em minus 0.4em\relax PMLR, 30 Oct--01 Nov 2020, pp. 86--99. [Online]. Available: \url{https://proceedings.mlr.press/v100/chai20a.html}
\BIBentrySTDinterwordspacing

\bibitem{Yue2022-os}
J.~Yue, D.~Manocha, and H.~Wang, ``Human trajectory prediction via neural social physics,'' July 2022.

\bibitem{Zhao2021-ca}
H.~Zhao and R.~P. Wildes, ``Where are you heading? dynamic trajectory prediction with expert goal examples,'' in \emph{2021 {IEEE/CVF} International Conference on Computer Vision ({ICCV})}.\hskip 1em plus 0.5em minus 0.4em\relax IEEE, Oct. 2021, pp. 7629--7638.

\bibitem{Yao2021-xn}
Y.~Yao, E.~Atkins, M.~Johnson-Roberson, R.~Vasudevan, and X.~Du, ``{BiTraP}: {Bi-Directional} pedestrian trajectory prediction with {Multi-Modal} goal estimation,'' \emph{IEEE Robotics and Automation Letters}, vol.~6, no.~2, pp. 1463--1470, Apr. 2021.

\bibitem{Wong2022-kx}
C.~Wong, B.~Xia, Z.~Hong, Q.~Peng, W.~Yuan, Q.~Cao, Y.~Yang, and X.~You, ``View vertically: A hierarchical network for trajectory prediction via fourier spectrums,'' in \emph{Computer Vision -- {ECCV} 2022}.\hskip 1em plus 0.5em minus 0.4em\relax Springer Nature Switzerland, 2022, pp. 682--700.

\bibitem{liu_energy-based_2021}
\BIBentryALTinterwordspacing
M.~Liu, T.~He, M.~Xu, and W.~Zhang, ``Energy-{Based} {Imitation} {Learning},'' \emph{arXiv:2004.09395 [cs, stat]}, Apr. 2021, arXiv: 2004.09395. [Online]. Available: \url{http://arxiv.org/abs/2004.09395}
\BIBentrySTDinterwordspacing

\bibitem{Xu2022-ki}
C.~Xu, W.~Mao, W.~Zhang, and S.~Chen, ``Remember intentions: {Retrospective-Memory-based} trajectory prediction,'' Mar. 2022.

\bibitem{liu_social_2021}
\BIBentryALTinterwordspacing
Y.~Liu, Q.~Yan, and A.~Alahi, ``Social {NCE}: {Contrastive} {Learning} of {Socially}-aware {Motion} {Representations},'' \emph{arXiv:2012.11717 [cs]}, Aug. 2021, arXiv: 2012.11717. [Online]. Available: \url{http://arxiv.org/abs/2012.11717}
\BIBentrySTDinterwordspacing

\bibitem{liu_towards_2021}
\BIBentryALTinterwordspacing
Y.~Liu, R.~Cadei, J.~Schweizer, S.~Bahmani, and A.~Alahi, ``Towards {Robust} and {Adaptive} {Motion} {Forecasting}: {A} {Causal} {Representation} {Perspective},'' \emph{arXiv:2111.14820 [cs]}, Nov. 2021, arXiv: 2111.14820. [Online]. Available: \url{http://arxiv.org/abs/2111.14820}
\BIBentrySTDinterwordspacing

\bibitem{mangalam_it_2020}
\BIBentryALTinterwordspacing
K.~Mangalam, H.~Girase, S.~Agarwal, K.-H. Lee, E.~Adeli, J.~Malik, and A.~Gaidon, ``It {Is} {Not} the {Journey} but the {Destination}: {Endpoint} {Conditioned} {Trajectory} {Prediction},'' \emph{arXiv:2004.02025 [cs]}, July 2020, arXiv: 2004.02025. [Online]. Available: \url{http://arxiv.org/abs/2004.02025}
\BIBentrySTDinterwordspacing

\bibitem{salzmann_trajectron_2021}
\BIBentryALTinterwordspacing
T.~Salzmann, B.~Ivanovic, P.~Chakravarty, and M.~Pavone, ``Trajectron++: {Dynamically}-{Feasible} {Trajectory} {Forecasting} {With} {Heterogeneous} {Data},'' \emph{arXiv:2001.03093 [cs]}, Jan. 2021, arXiv: 2001.03093. [Online]. Available: \url{http://arxiv.org/abs/2001.03093}
\BIBentrySTDinterwordspacing

\bibitem{kipf2018neural}
T.~Kipf, E.~Fetaya, K.-C. Wang, M.~Welling, and R.~Zemel, ``Neural relational inference for interacting systems,'' in \emph{International Conference on Machine Learning}.\hskip 1em plus 0.5em minus 0.4em\relax PMLR, 2018, pp. 2688--2697.

\bibitem{alet_neural_2019}
F.~Alet, E.~Weng, T.~Lozano-Pérez, and L.~P. Kaelbling, ``\BIBforeignlanguage{en}{Neural {Relational} {Inference} with {Fast} {Modular} {Meta}-learning},'' p.~12.

\bibitem{Liang2020-of}
M.~Liang, B.~Yang, R.~Hu, Y.~Chen, R.~Liao, S.~Feng, and R.~Urtasun, ``Learning lane graph representations for motion forecasting,'' July 2020.

\bibitem{chen_relational_2020}
\BIBentryALTinterwordspacing
C.~Chen, S.~Hu, P.~Nikdel, G.~Mori, and M.~Savva, ``Relational {Graph} {Learning} for {Crowd} {Navigation},'' \emph{arXiv:1909.13165 [cs]}, Aug. 2020, arXiv: 1909.13165. [Online]. Available: \url{http://arxiv.org/abs/1909.13165}
\BIBentrySTDinterwordspacing

\bibitem{su_trajectory_2022}
\BIBentryALTinterwordspacing
Y.~Su, J.~Du, Y.~Li, X.~Li, R.~Liang, Z.~Hua, and J.~Zhou, ``Trajectory {Forecasting} {Based} on {Prior}-{Aware} {Directed} {Graph} {Convolutional} {Neural} {Network},'' \emph{IEEE Transactions on Intelligent Transportation Systems}, pp. 1--13, 2022, conference Name: IEEE Transactions on Intelligent Transportation Systems. [Online]. Available: \url{10.1109/TITS.2022.3142248}
\BIBentrySTDinterwordspacing

\bibitem{mohamed_social-stgcnn_2020}
\BIBentryALTinterwordspacing
A.~Mohamed, K.~Qian, M.~Elhoseiny, and C.~Claudel, ``Social-{STGCNN}: {A} {Social} {Spatio}-{Temporal} {Graph} {Convolutional} {Neural} {Network} for {Human} {Trajectory} {Prediction},'' \emph{arXiv:2002.11927 [cs]}, Mar. 2020, arXiv: 2002.11927 version: 3. [Online]. Available: \url{http://arxiv.org/abs/2002.11927}
\BIBentrySTDinterwordspacing

\bibitem{Yu2020-df}
C.~Yu, X.~Ma, J.~Ren, H.~Zhao, and S.~Yi, ``{Spatio-Temporal} graph transformer networks for pedestrian trajectory prediction,'' in \emph{Computer Vision -- {ECCV} 2020}.\hskip 1em plus 0.5em minus 0.4em\relax Springer International Publishing, 2020, pp. 507--523.

\bibitem{Yuan2020-vw}
Y.~Yuan and K.~Kitani, ``{DLow}: Diversifying latent flows for diverse human motion prediction,'' Mar. 2020.

\bibitem{Yuan2021-tp}
Y.~Yuan, X.~Weng, Y.~Ou, and K.~Kitani, ``{AgentFormer}: {Agent-Aware} transformers for {Socio-Temporal} {Multi-Agent} forecasting,'' pp. 9813--9823, Mar. 2021.

\bibitem{Amirloo2022-se}
E.~Amirloo, A.~Rasouli, P.~Lakner, M.~Rohani, and J.~Luo, ``{LatentFormer}: {Multi-Agent} {Transformer-Based} interaction modeling and trajectory prediction,'' Mar. 2022.

\bibitem{Cao2022-uv}
Z.~Cao, E.~B{\i}y{\i}k, G.~Rosman, and D.~Sadigh, ``Leveraging smooth attention prior for {Multi-Agent} trajectory prediction,'' Mar. 2022.

\bibitem{bahari_svg-net_2021}
\BIBentryALTinterwordspacing
M.~Bahari, V.~Zehtab, S.~Khorasani, S.~Ayromlou, S.~Saadatnejad, and A.~Alahi, ``{SVG}-{Net}: {An} {SVG}-based {Trajectory} {Prediction} {Model},'' \emph{arXiv:2110.03706 [cs]}, Oct. 2021, arXiv: 2110.03706. [Online]. Available: \url{http://arxiv.org/abs/2110.03706}
\BIBentrySTDinterwordspacing

\bibitem{Huang2021-sv}
Z.~Huang, X.~Mo, and C.~Lv, ``Multi-modal motion prediction with transformer-based neural network for autonomous driving,'' Sept. 2021.

\bibitem{c10}
\BIBentryALTinterwordspacing
H.~Caesar, V.~Bankiti, A.~H. Lang, S.~Vora, V.~E. Liong, Q.~Xu, A.~Krishnan, Y.~Pan, G.~Baldan, and O.~Beijbom, ``nuscenes: {A} multimodal dataset for autonomous driving,'' \emph{CoRR}, vol. abs/1903.11027, 2019. [Online]. Available: \url{http://arxiv.org/abs/1903.11027}
\BIBentrySTDinterwordspacing

\bibitem{womo}
\BIBentryALTinterwordspacing
S.~Ettinger, S.~Cheng, B.~Caine, C.~Liu, H.~Zhao, S.~Pradhan, Y.~Chai, B.~Sapp, C.~R. Qi, Y.~Zhou, Z.~Yang, A.~Chouard, P.~Sun, J.~Ngiam, V.~Vasudevan, A.~McCauley, J.~Shlens, and D.~Anguelov, ``Large scale interactive motion forecasting for autonomous driving : The waymo open motion dataset,'' \emph{CoRR}, vol. abs/2104.10133, 2021. [Online]. Available: \url{https://arxiv.org/abs/2104.10133}
\BIBentrySTDinterwordspacing

\bibitem{chang_argoverse_2019}
\BIBentryALTinterwordspacing
M.-F. Chang, D.~Ramanan, J.~Hays, J.~Lambert, P.~Sangkloy, J.~Singh, S.~Bak, A.~Hartnett, D.~Wang, P.~Carr, and S.~Lucey, ``\BIBforeignlanguage{en}{Argoverse: {3D} {Tracking} and {Forecasting} {With} {Rich} {Maps}},'' in \emph{\BIBforeignlanguage{en}{2019 {IEEE}/{CVF} {Conference} on {Computer} {Vision} and {Pattern} {Recognition} ({CVPR})}}.\hskip 1em plus 0.5em minus 0.4em\relax Long Beach, CA, USA: IEEE, June 2019, pp. 8740--8749. [Online]. Available: \url{https://ieeexplore.ieee.org/document/8953693/}
\BIBentrySTDinterwordspacing

\bibitem{Geiger2013-xv}
A.~Geiger, P.~Lenz, C.~Stiller, and R.~Urtasun, ``\BIBforeignlanguage{en}{Vision meets robotics: The {KITTI} dataset},'' \emph{\BIBforeignlanguage{en}{Int. J. Rob. Res.}}, vol.~32, no.~11, pp. 1231--1237, Sept. 2013.

\bibitem{c12}
\BIBentryALTinterwordspacing
J.~Liang, L.~Jiang, and A.~G. Hauptmann, ``Simaug: Learning robust representations from 3d simulation for pedestrian trajectory prediction in unseen cameras,'' \emph{CoRR}, vol. abs/2004.02022, 2020. [Online]. Available: \url{https://arxiv.org/abs/2004.02022}
\BIBentrySTDinterwordspacing

\bibitem{liang_garden_2020-1}
\BIBentryALTinterwordspacing
J.~Liang, L.~Jiang, K.~Murphy, T.~Yu, and A.~Hauptmann, ``The {Garden} of {Forking} {Paths}: {Towards} {Multi}-{Future} {Trajectory} {Prediction},'' \emph{arXiv:1912.06445 [cs]}, Mar. 2020, arXiv: 1912.06445. [Online]. Available: \url{http://arxiv.org/abs/1912.06445}
\BIBentrySTDinterwordspacing

\bibitem{biswas_socnavbench_2021}
\BIBentryALTinterwordspacing
A.~Biswas, A.~Wang, G.~Silvera, A.~Steinfeld, and H.~Admoni, ``{SocNavBench}: {A} {Grounded} {Simulation} {Testing} {Framework} for {Evaluating} {Social} {Navigation},'' \emph{arXiv:2103.00047 [cs]}, July 2021, arXiv: 2103.00047. [Online]. Available: \url{http://arxiv.org/abs/2103.00047}
\BIBentrySTDinterwordspacing

\bibitem{kothari_human_2021}
\BIBentryALTinterwordspacing
P.~Kothari, S.~Kreiss, and A.~Alahi, ``Human {Trajectory} {Forecasting} in {Crowds}: {A} {Deep} {Learning} {Perspective},'' \emph{arXiv:2007.03639 [cs]}, Jan. 2021, arXiv: 2007.03639. [Online]. Available: \url{http://arxiv.org/abs/2007.03639}
\BIBentrySTDinterwordspacing

\bibitem{Van_den_Berg2011-yv}
J.~van~den Berg, S.~J. Guy, M.~Lin, and D.~Manocha, ``Reciprocal n-body collision avoidance,'' in \emph{Springer Tracts in Advanced Robotics}, ser. Springer tracts in advanced robotics.\hskip 1em plus 0.5em minus 0.4em\relax Berlin, Heidelberg: Springer Berlin Heidelberg, 2011, pp. 3--19.

\bibitem{Guo2021-ec}
K.~Guo, D.~Wang, T.~Fan, and J.~Pan, ``{VR-ORCA}: Variable responsibility optimal reciprocal collision avoidance,'' \emph{IEEE Robotics and Automation Letters}, vol.~6, no.~3, pp. 4520--4527, July 2021.

\bibitem{Chen2016-xm}
Y.~F. Chen, M.~Liu, M.~Everett, and J.~P. How, ``Decentralized non-communicating multiagent collision avoidance with deep reinforcement learning,'' Sept. 2016.

\bibitem{Liang2019-xo}
J.~Liang, L.~Jiang, K.~Murphy, T.~Yu, and A.~Hauptmann, ``The garden of forking paths: Towards {Multi-Future} trajectory prediction,'' Dec. 2019.

\bibitem{Silvera2022-uk}
G.~Silvera, A.~Biswas, and H.~Admoni, ``{DReyeVR}: Democratizing virtual reality driving simulation for behavioural \& interaction research,'' Jan. 2022.

\bibitem{c16}
F.~Camara, P.~Dickinson, N.~Merat, and C.~Fox, ``Examining pedestrian-autonomous vehicle interactions in virtual reality,'' 09 2019.

\bibitem{yuan_agentformer_2021}
\BIBentryALTinterwordspacing
Y.~Yuan, X.~Weng, Y.~Ou, and K.~Kitani, ``{AgentFormer}: {Agent}-{Aware} {Transformers} for {Socio}-{Temporal} {Multi}-{Agent} {Forecasting},'' \emph{arXiv:2103.14023 [cs]}, Oct. 2021, arXiv: 2103.14023. [Online]. Available: \url{http://arxiv.org/abs/2103.14023}
\BIBentrySTDinterwordspacing

\bibitem{c5}
\BIBentryALTinterwordspacing
A.~Kalatian and B.~Farooq, ``A context-aware pedestrian trajectory prediction framework for automated vehicles,'' \emph{CoRR}, vol. abs/2104.08123, 2021. [Online]. Available: \url{https://arxiv.org/abs/2104.08123}
\BIBentrySTDinterwordspacing

\bibitem{c17}
E.~Weng, H.~Hoshino, D.~Ramanan, and K.~Kitani, ``Joint metrics matter: A better standard for trajectory forecasting,'' 2023.

\end{thebibliography}

\end{document}